\documentclass{esannV2}
\usepackage[latin1]{inputenc}
\usepackage{amssymb,amsmath,array, bm}
\usepackage{multirow} 
\usepackage{booktabs} 
\usepackage{times} 
\usepackage{graphicx} 
\usepackage{dblfloatfix} 

\DeclareMathOperator*{\argmin}{arg\,min}

\begin{document}
\title{Behavior Constraining in Weight Space\\ for Offline Reinforcement Learning}

\author{Phillip Swazinna$^{12}$, Steffen Udluft$^1$, Daniel Hein$^1$ and Thomas Runkler $^{12}$
\thanks{The project this paper is based on was supported with funds from the German Federal Ministry of Education and Research under project number 01IS18049A.}
%
\vspace{.3cm}\\
%
1) Siemens Technology - Learning Systems - Germany 
%
\vspace{.1cm}\\
2) Technical University of Munich - Dept. of Informatics - Germany 
}

\maketitle

\begin{abstract}
In offline reinforcement learning, a policy needs to be learned from a single pre-collected dataset. Typically, policies are thus regularized during training to behave similarly to the data generating policy, by adding a penalty based on a divergence between action distributions of generating and trained policy. We propose a new algorithm, which constrains the policy directly in its weight space instead, and demonstrate its effectiveness in experiments.
\end{abstract}

\section{Introduction \& Related Work}
Reinforcement learning (RL) algorithms \cite{intro} have demonstrated success in a variety of tasks, be it traditional board games, video game control, or robotic locomotion \cite{rl_overview}. Most algorithms rely on directly interacting with the environment in which they should perform and switch back and forth between training a policy and using it to collect further data. In batch RL, algorithms are constrained to work with a single dataset of transitions obtained from the system, which is much closer to how practitioners can use RL in real-world control tasks due to safety concerns. However, most works still make unrealistic assumptions about how the dataset was obtained - datasets are often collected by uniform random policies, ensuring that the data reflects the entire transition dynamics of the system well. Only very recently, scientific literature has been concerned with deriving policies from datasets that have been collected under sub-optimal exploration settings, moving closer to real-world problems where large datasets have often been collected under a few operational policies already in existence. This setting is commonly referred to as offline RL \cite{bcq, bear, brac, moose}.\\
While other works in this area usually train an additional generative model to mimic the behavior policies and then impose a penalty based on some divergence measure between the trained and behavior policy in action space during training, we will in the following present a much simpler, population-based learning algorithm, which constrains the behavior directly in policy weight space. After developing our approach, we show empirically that it performs well compared to other offline RL algorithms.

\section{The Industrial Benchmark}
As the motivation for the general offline RL problem class is derived from realistic assumptions about real-world control problems, we also choose the benchmark environment to reflect problems often encountered in such tasks: The industrial benchmark (IB) \cite{IB} has been developed to include many aspects commonly found in industrial control problems, such as high-dimensional state and action spaces which are only partially observable, heteroscedastic transition noise, delayed rewards, and multiple opposing reward components that need trade-off. While not being a simulation of a specific industrial system, the IB poses similar hardness and complexity as most of these environments do. An open source implementation is available online\footnote{https://github.com/siemens/industrialbenchmark}.\\
The observable part of the state space is in each time step comprised of the three steerings velocity $v$, gain $g$, and shift $h$, the setpoint $p$, and the two reward components fatigue $f$ and consumption $c$. The actions given by the policy can be interpreted as proposed changes in the three steerings, i.e., $a_t = (\Delta v, \Delta g, \Delta h)$. While the steerings are updated in a deterministic fashion based on the actions, the reward components are influenced by multiple, complex, stochastic sub-dynamics, which are invisible to the agent. For a detailed mathematical description of the sub-dynamics, we refer to \cite{IB}. As the reward components are influenced not only by the observable state at some fixed time, but also by past states, the effective state space that a policy or transition model has to consider, grows. While the actual, observable state space contains only six variables per time step, usually the 30 most recent time steps are combined to form a 180-dimensional state space on which models and policies are trained. The reward is computed as $r_t = -c_t -3f_t$.

\section{Weight Space Behavior Constraining}
In the offline RL setting, we have to take into account that the method used to estimate policy performance might not be fully trustworthy. Instead, we assume that the performance estimate can only be true for policies which are by some metric close to the policies that generated the dataset, since otherwise corresponding transitions have not been seen in the data and can thus not be predicted reliably. While most offline RL methods are model-free (e.g. \cite{bcq, bear, brac}), recent works have shown that using a transition model can improve performance as well as increase the stability of learning \cite{moose}. Since model-based approaches are regarded to be more data-efficient than model-free ones \cite{moose}, it seems straightforward that they have an advantage in the offline RL setting since it is innately data-limited. Most published offline RL methods have trained additional representations of the behavior policies that generated the data and use it during training to compare action distributions of original and trained policies in order to penalize divergences. In this work we take the novel approach of constraining the behavior of the trained policy in weight space instead. In the following, we derive our algorithm, Weight Space Behavior Constraining (WSBC), and show experimentally that it is not only much simpler, but also that it can be much more efficient to constrain the policies in this way, and that it outperforms previous algorithms across a variety of datasets.

\subsection{Transition Models}
We train neural models $f$ parameterized by $\phi$ of the transition dynamics based on the dataset $\mathrm{D}$ containing $N=100{,}000$ transitions: $\mathrm{D} = \{(s_i, a_i, s_{i+1})\}_{i=0}^{N-1}$. Due to the delayed reward effects, we employ recurrent networks (single recurrent cell with $30$ units and $\rm{tanh}$ nonlinearity, followed by a linear layer). During training, we use the $H_p=30$ past time step states to construct the hidden state $h$ of the model and then predict the next $H_f=50$ future steps without the real states from the dataset, but instead with its own state predictions and the actions of the corresponding time steps in the dataset. This method of overshooting instead of only predicting the single next time step helps the model to predict long rollouts more accurately, since it learns to use its own prediction of the current step for future predictions as well. The models are trained via gradient descent on a maximum likelihood objective:
\begin{align}
    &L(\phi) = \mathbb{E}_{s_0 \sim D} \left[ \sum_{t=H_p}^{H_p+H_f} [g(f_{\phi}(\hat{s}_t, a_t, h_{t-1})) - s_{t+1}]^2 \right]\\
    \rm{with} \quad 
    \hat{s}_{t+1} = &\begin{cases}
      s_{t+1} &\text{if} \quad t < H_p\\
      g(f_{\phi}(\hat{s}_t, a_t, h_{t-1})) &\text{else}\\
    \end{cases}
    \quad \mathrm{and} \quad
    h_{t} = h(f_{\phi}(\hat{s}_t, a_t, h_{t-1})) \nonumber
\end{align}
where $\hat{s}_0 = s_0$, $h_0=\bm{\Vec{0}}$, and $g(f(\cdot))$ and $h(f(\cdot))$ extract the state prediction and the new hidden state from the output of the model $f$. We train $K=4$ transition models.

\subsection{Policy Search}
We aim to find neural policies $\pi$ parameterized by $\theta$, which will assign the best possible action $a_t$ given the current state history $s_t,...,s_{t-H_p}$ in a deterministic fashion: $a_t = \pi_{\theta}(s_t,...,s_{t-H_p})$. The policies have a single hidden layer of size $20$ with $\rm{ReLU}$ nonlinearity. We estimate their performance by conducting rollouts of the policies through the trained transition model ensemble. Similarly to the model training, we use $H_p=30$ past states to build up the hidden state and then predict $H_f=100$ steps into the future. However, this time the actions are not determined by the dataset, but instead by the policy candidate based on the state predictions of the model. We obtain a conservative estimate of the return by taking the minimum over members in the ensemble in each step:
\begin{align}
    L(\theta) &= \mathbb{E}_{\pi}[R] = \frac{1}{|S|} \sum_{s_0 \in S} \sum_{t=H_p}^{H_p+H_f} \gamma^t r(g(f_{\phi}^{k_t}(\hat{s}_t, \pi_{\theta}(\hat{s}_t..\hat{s}_{t-H_p}), h_{t-1})))\\
    \rm{s.t.} \quad k_t &= \argmin_k r(g(f_{\phi}^k(\hat{s}_t, \pi_{\theta}(\hat{s}_t..\hat{s}_{t-H_p}), h_{t-1}))) \quad k \in \{0,...,K-1\} \nonumber \\
    \rm{with} \quad 
    \hat{s}_{t+1} &= \begin{cases}
      g(f_{\phi}(\hat{s}_t, a_t, h_{t-1})) &\text{if} \quad t < H_p\\
      g(f_{\phi}(\hat{s}_t, \pi_{\theta}(\hat{s}_t..\hat{s}_{t-H_p}), h_{t-1})) &\text{else}\\
    \end{cases} \nonumber \\
    \rm{and} \quad
    h_{t} &= \begin{cases}
      h(f_{\phi}(\hat{s}_t, a_t, h_{t-1})) &\text{if} \quad t < H_p\\
      h(f_{\phi}(\hat{s}_t, \pi_{\theta}(\hat{s}_t..\hat{s}_{t-H_p}), h_{t-1})) &\text{else.}\\
    \end{cases} \nonumber
\end{align}
Since we do not trust our model-ensemble to predict the policy performance with a reasonable degree of accuracy for policies that deviate strongly from those that generated the data, we need to restrict the policy search to only include policies that behave similarly to the generating policies. A common approach used in this context is to augment the loss by a term that penalizes divergence from the behavior policy $\beta$ in the action distribution:
\begin{equation}
    L(\theta) = \mathbb{E}_{\pi}[R] - \alpha \mathbb{E}[D(\pi_{\theta}(a|s) || \beta(a|s))].
\end{equation}
Instances of such algorithms are for example BRAC or BEAR \cite{bear, brac}, which estimate $\mathbb{E}[R]$ with a Q-function and use the KL divergence or the maximum mean discrepancy as $D$, respectively. We will however constrain the policies strictly in their weight space, by clipping weights $\theta$ that are further away than $d$ by some norm $||\cdot||$ from the weights $\psi$ of the behavior policy $\beta_{\psi}$, which is trained directly from the dataset:
\begin{align}
    L(\theta) &= \mathbb{E}_{\pi}[R] \quad \rm{s.t.} \quad \forall i: ||\theta_i - \psi_i|| < d.\\
    L(\psi) &= \frac{1}{N} \sum_{t} [\beta_{\psi}(s_t,...,s_{t-H_p}) - a_t]^2 \nonumber
\end{align}
The chosen model-rollout based approach of estimating policy performance has been shown to increase data efficiency and lower the variance in the trained policies \cite{moose}. However in long horizon rollouts, a gradient based policy search can be cumbersome: Since returns are estimated based on entire trajectories instead of individual state-action pairs (as in value functions), the credit assignment to individual actions is much harder. When the gradient is not only computed with respect to the action samples, but rather computed backwards throughout the model, we can also be faced with exploding or vanishing gradients due to the long horizon. To avoid these issues altogether, we search policies by particle swarm optimization (PSO) \cite{pso}, which doesn't require any gradients and makes few assumptions about the optimization problem. PSO instead uses neighborhood information to systematically search for well performing regions in a potentially very large policy space. It has thus been well established as a method for learning neural policies in a model-rollout context \cite{pso_fuzzy}. Each particle's ($=$policy's) position $P_n^i$ at iteration $i$ is represented by its weight vector. Every particle also has a velocity $V_n^i$. Both are updated via:
\begin{align}
    P_n^{i+1} &= P_n^i + V_n^i \\
    V_n^{i+1} &= wV_n^i + c_1r_1(P_{\rm{best}(n)}^i - P_n^i) + c_2r_2(P_{\rm{bestneighbor}(n)}^i - P_n^i) \nonumber
\end{align}
where $c_1$ and $c_2$ are constants, $r_1$ and $r_2$ are random vectors, $P_{\rm{best}(n)}^i$ is the best position in which particle $n$ has been up to iteration $i$, and $P_{\rm{bestneighbor}(n)}^i$ is the best position that $n$'s neighborhood has seen up to iteration $i$. During policy search we use $200$ particles and a ring neighborhood \cite{pso_fuzzy} of size $30$ to balance global and local search. The rollouts are conducted with $\gamma=0.97$ and with the L1 norm to constrain policy weights.

\section{Experiments}
We perform experiments on 16 datasets collected on the IB with different behavior policies to ensure broad applicability of our algorithm and compare performance with different model-free and model-based offline RL algorithms. We report tenth percentile performances in Table \ref{table:performance}, since we cannot know \textit{a priori} which policies will perform well. The datasets are collected with the same three deterministic baseline policies as \cite{moose}: $\{\rm{bad, mediocre, optimized}\}$, each augmented with six different degrees of $\varepsilon$-greedy exploration $\varepsilon=\{0.0, 0.2, 0.4, 0.6, 0.8, 1.0\}$. The bad and mediocre policies aim to drive the system to a fixed point in the steering space, which is characteristic for controllers in many industrial systems. The optimized policy is more complex, however still interpretable, and constitutes a realistic example of an expert-crafted policy:
\begin{equation}
    \pi_{\rm{bad}} = \begin{cases} 100 - v_t\\ 100 - g_t\\ 100 - h_t \end{cases}
    \pi_{\rm{med}} = \begin{cases} 25 - v_t\\ 25 - g_t\\ 25 - h_t \end{cases}
    \pi_{\rm{opt}} = \begin{cases} \hfil -v_{t-5} - 0.91\\ \hfil 2 f_{t-3} - p + 1.43\\
    -3.48 h_{t-3} - h_{t-4} + 2 p + 0.81 \end{cases} \nonumber
\end{equation}

\setlength{\tabcolsep}{1.15pt}
\begin{table*}
\caption{\label{table:performance} Tenth percentile performances and their standard error}
  \begin{center}
    \scriptsize
    \makebox[0pt]{\begin{tabular}{|ccc|ccc||ccrlcc|ccrlcc|ccrlcc|ccrlcc|ccrlcc|ccrlcc|} 
    \toprule
       \multicolumn{6}{|c||}{$\varepsilon = $} & \multicolumn{6}{c|}{0.0} & \multicolumn{6}{c|}{0.2} & \multicolumn{6}{c|}{0.4} & \multicolumn{6}{c|}{0.6} & \multicolumn{6}{c|}{0.8} & \multicolumn{6}{c|}{1.0} \\
      \midrule
      & \parbox[t]{2mm}{\multirow{5}{*}{\rotatebox[origin=c]{90}{Bad}}} &
     & & BRAC-v & & & & -274 & (12) & & & & & -270.2 & (12) & & & & & -199 & (7) & & & & & -188 & (8) & & & & & -140 & (5) & & & & & & & & \\
     & & & & BEAR & & & & -322 & (4) & & & & & -168 & (5) & & & & & -129 & (4) & & & & & -90& (1) & & & & & -90 & (1) & & & & & & & & \\
      & & & & BCQ & & & & -313 & (1) & & & & & -281 & (3) & & & & & -234 & (5) & & & & & -127 & (4) & & & & & -89 & (2) & & & & & & & & \\
      & & & & MOOSE & & & & -311 & (1) & & & & & -128& (1) & & & & & -110& (1) & & & & & -92.7 & (0.4) & & & & & -71.3 & (0.2) & & & & & & & &  \\
      & & & & ours & & & & \textbf{-134} & \textbf{(2)} & & & & & \textbf{-118}&\textbf{(1)} & & & & & \textbf{-103}& \textbf{(1)} & & & & & \textbf{-84.9} &\textbf{(0.2)} & & & & & \textbf{-70.0} & \textbf{(0.1)} & & & & & & & & \\

     \cline{4-36}
      &\parbox[t]{2mm}{\multirow{5}{*}{\rotatebox[origin=c]{90}{Mediocre}}} &
     & & BRAC-v & & & & -117 & (3) & & & & & -98.3 & (2.3) & & & & & -90.8 & (1) & & & & & -91.3 & (4.8) & & & & & -95.3 & (2.5) & & & & & -113 & (3) & & \\
     & & & & BEAR & & & & -111 & (1) & & & & & -115 & (7) & & & & & -109 & (4) & & & & & -111 & (6) & & & & & -104 & (3) & & & & & -65.1 & (0.3) & & \\
      & & & & BCQ & & & & -105 & (2) & & & & & -77.1 & (0.2) & & & & & -71.2& (0.3) & & & & & -78 & (1) & & & & & -125 & (4) & & & & & -68.6 & (0.3) & &\\
     & & & & MOOSE & & & & -83.3& (0.3) & & & & & -76.6& (0.1) & & & & & -75.0 & (0.1) & & & & & \textbf{-71.1}& \textbf{(0.1)} & & & & & -69.7& (0.4) & & & & & -64.11& (0.02) & & \\
      & & & & ours & & & & \textbf{-71.1}&\textbf{(0.2)} & & & & & \textbf{-68.5}&\textbf{(0.1)} & & & & & \textbf{-68.9} & \textbf{(0.1)} & & & & & -243& (1) & & & & & \textbf{-62.9}&\textbf{(0.2)} & & & & & \textbf{-63.76}&\textbf{(0.04)} & & \\
     
     \cline{4-36}
      &\parbox[t]{2mm}{\multirow{5}{*}{\rotatebox[origin=c]{90}{Optimized}}} &
     & & BRAC-v & & & & -127 & (5) & & & & & -78.4 & (0.9) & & & & & -165 & (6) & & & & & -76.9 & (2.1) & & & & & -98.7 & (3.1) & & & & & & & & \\
      & & & & BEAR & & & & -60.5 & (0.6) & & & & & -61.7 & (0.1) & & & & & -64.7 & (0.3) & & & & & -64.3 & (0.2) & & & & & -63.1 & (0.2) &  & & & & & & & \\
      & & & & BCQ & & & & -60.1 & (0.1) & & & & & -60.6 & (0.1) & & & & & -62.4 & (0.2) & & & & & -62.7 & (0.2) & & & & & -74.1 & (0.8) &  & & & & & & & \\
      & & & & MOOSE & & & & \textbf{-59.76}&\textbf{(0.02)} & & & & & -60.35& (0.02) & & & & & -60.77& (0.02) & & & & & -62.04& (0.02) & & & & & -62.73& (0.03) &  & & & & & & & \\
      & & & & ours & & & & -60.18& (0.03) & & & & & \textbf{-58.2}& \textbf{(0.1)} & & & & & \textbf{-58.6}&\textbf{(0.1)} & & & & & \textbf{-59.39}&\textbf{(0.02)} & & & & & \textbf{-61.70}& \textbf{(0.01)} &  & & & & & & & \\
     \bottomrule
    \end{tabular}}
  \end{center}
\end{table*}

\subsection{Alternatives \& Ablation}
\begin{figure}
    \centering
    \includegraphics[width=\textwidth]{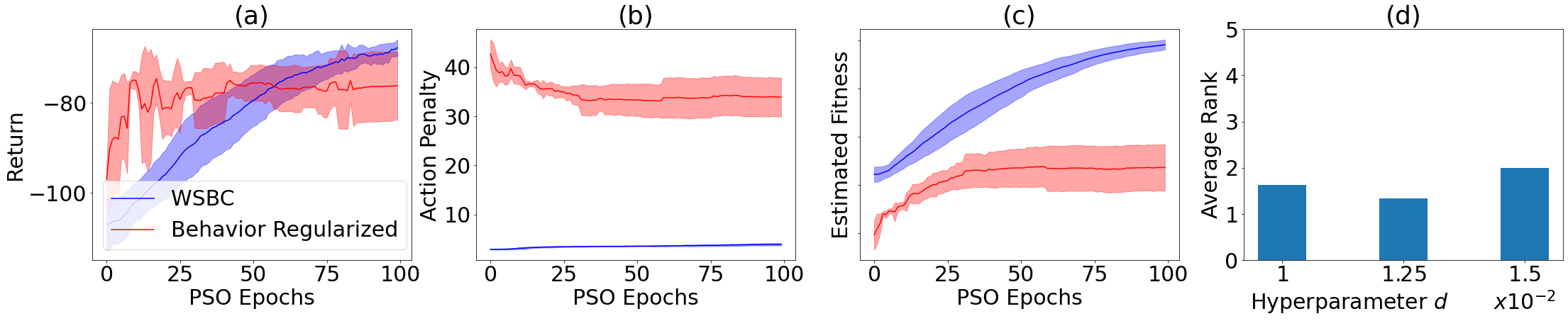}
    \caption{(a-c) Show return, action space penalty, and fitness ($=$ estimated return $- \alpha \times$ action space penalty) of WSBC and explicit action space behavior regularization as defined by Eq. \ref{eq:AP} on bad-$0.8$. WSBC achieves lower action space penalty and better fitness values in terms of Eq. \ref{eq:AP} without explicitly optimizing for it. (d) Average rank of WSBC across datasets when the hyperparameter $d$ was altered.}
    \label{fig:compare}
\end{figure}

We briefly compare the proposed approach with a direct penalty in action space which is factored into the loss function: Similarly to MOOSE \cite{moose}, we train a variational autoencoder $\beta(\cdot|s)$ to obtain a policy model from the dataset $\mathrm{D}$ and use it to derive a penalty based on the mean squared reconstruction error, which we use instead of the direct weight space constraining:
\begin{equation}
\label{eq:AP}
    L(\theta) = \mathbb{E}_{\pi}[R] - \alpha \mathbb{E}[(\pi_{\theta}(s) - \beta(a|s))^2].
\end{equation}
We scale $\alpha$ so that in the initial PSO population $\alpha \mathbb{E}[(\pi_{\theta}(s) - \beta(a|s))^2] = 0.5 |\mathbb{E}_{\pi}[R]|$. In Fig. \ref{fig:compare} (a-c), we report return, action space penalty, and fitness of both approaches, and find that our approach produces policies with lower action space penalty and consequently better combined fitness (as by Eq. \ref{eq:AP}) and return than the direct action space constrained policy search.\\
Furthermore, we alter the parameter $d$ and redo the experiments on all datasets to investigate the sensitivity of the algorithm with respect to the hyperparameter. We calculate the average rank compared to MOOSE, BCQ, BEAR, and BRAC-v \cite{bcq, bear, brac, moose}, and report results in Figure \ref{fig:compare} (d).

\section{Conclusion}
In this paper, we presented a novel approach to offline RL: Instead of the plethora of ways in which a penalty based on the action distribution of a policy can be derived, we constrain the policy directly in its weight space. Even though our method doesn't explicitly constrain equality in the actions output by a policy, our experiments show that it does so implicitly, and that the effect can be much stronger than explicit behavior regularization. We compare our algorithm on the IB against other state of the art offline RL algorithms and find that we can outperform most of them reliably, even though the performance naturally depends on the chosen hyperparameter $d$, which is however a common problem all of the offline RL methods share.


\begin{footnotesize}



\end{footnotesize}


\end{document}